\pdfoutput=1
\documentclass[11pt]{article}

\usepackage{acl}

\usepackage{times}
\usepackage{latexsym}

\usepackage[T1]{fontenc}
\usepackage[utf8]{inputenc}

\usepackage{microtype}

\usepackage{inconsolata}

\usepackage{natbib}
\usepackage{multibib}
\usepackage{graphicx}
\usepackage{tabularx}
\usepackage{soul}
\usepackage{titlesec}

\usepackage{xcolor}
\usepackage{hyperref}
 \definecolor{darkblue}{rgb}{0, 0, 0.5}
  \hypersetup{colorlinks=true, citecolor=darkblue, linkcolor=darkblue, urlcolor=darkblue}

\usepackage{xstring}
\usepackage{color}

\usepackage{capt-of}
\usepackage{blindtext}

\usepackage{paralist}

\usepackage{graphicx}
\usepackage{amsmath}

\usepackage[normalem]{ulem}
\useunder{\uline}{\ul}{}

\usepackage{array,multirow,graphicx}

\title{(Chat)GPT \includegraphics[scale=0.04]{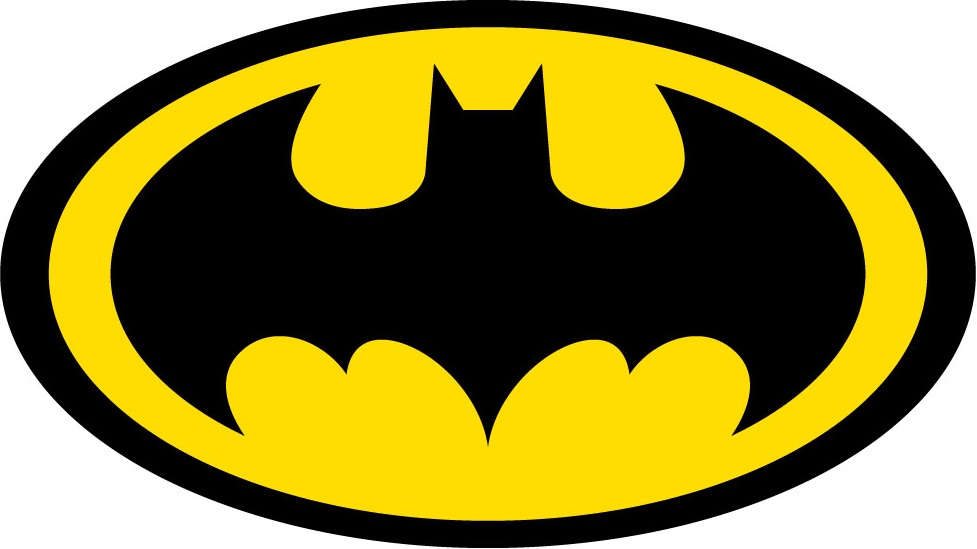} v BERT \includegraphics[scale=0.013]{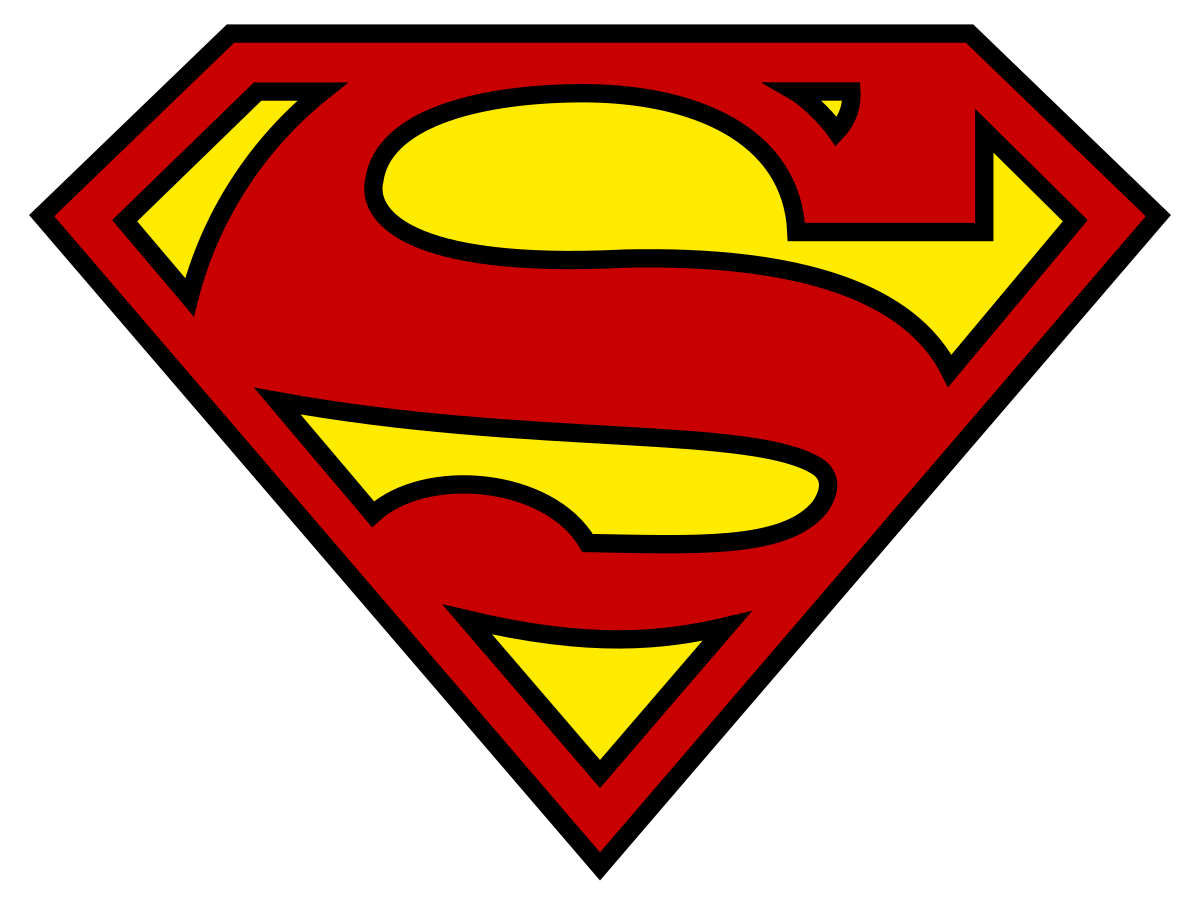}\\ Dawn of Justice for \textit{Semantic Change Detection}
\\{\small \textbf{\url{https://aclanthology.org/2024.findings-eacl.29.pdf}}}}

\author{Francesco Periti \ \ \\
  University of Milan \ \ \\
  Via Celoria 18 \ \ \\
  20133 Milan, Italy \ \ \\
 {\small \texttt{francesco.periti@unimi.it}} \\\And
  \ Haim Dubossarsky\\
  \ Queen Mary University of London \ \ \\
  \ Mile End Road\\
  \ E1 4NS London, United Kingdom\ \ \\
  {\small \texttt{h.dubossarsky@qmul.ac.uk}} \\ \And
  Nina Tahmasebi\\
  \ \ University of Gothenburg\\
  Renstr{\"o}msgatan 6\\
  \ \ 40530 G{\"o}teborg, Sweden\\
  {\small \texttt{nina.tahmasebi@gu.se}} \\}

\begin{document}
\maketitle

\begin{abstract}
In the \textit{universe} of Natural Language Processing, Transformer-based language models like BERT and (Chat)GPT have emerged as \textit{lexical superheroes} with \textit{great power} to solve open research problems. In this paper, we specifically focus on the temporal problem of semantic change, and evaluate their ability to solve two diachronic extensions of the Word-in-Context (WiC) task: TempoWiC and HistoWiC. In particular, we investigate the potential of a novel, off-the-shelf technology like ChatGPT (and GPT) 3.5 compared to BERT, which represents a family of models that currently stand as the state-of-the-art for modeling semantic change. Our experiments represent the first attempt to assess the use of (Chat)GPT for studying semantic change. Our results indicate that ChatGPT performs significantly worse than the foundational GPT version. Furthermore, our results demonstrate that (Chat)GPT achieves slightly lower performance than BERT in detecting long-term changes but performs significantly worse in detecting short-term changes.
\end{abstract}

\section{Introduction}
Lexical semantic change is the linguistic phenomenon that denotes words changing their meanings over time~\cite{geeraerts2024lexical,bloomfield1933language}. An example is the word \texttt{gay} that changed from meaning \texttt{cheerful} to \texttt{homosexual} in the last century. This change is crucial to our understanding of historical texts. A nuanced grasp of semantic \textit{variation} between groups and genre, and semantic \textit{change} across time allows us to study languages, cultures, and societies through digitized text and opens up a range of research applications. Computational
approaches to semantic change are thus tools with immense potential for a range of research fields~\cite{montanelli2023survey,nina2021survey,kutuzov2018diachronic,tang2018state}. Not only can they broaden the field of historical linguistics and simplify lexicography, but they can also be fruitfully applied in the fields of sociology, history, and other text-based research. For instance, the computational modeling of semantic change is equally relevant when studying out-of-domain texts where language differs from the general language, like in medical~\cite{kay1979lexemic} and olfactory~\cite{paccosi2023sensibility,menini2022multilingual} domains. %legal texts or in radical groups.

The recent introduction of Transformer-based~\cite{Vaswani2017attention} language models (LMs) has led to significant advances in Natural Language Processing (NLP). These advances are exemplified in Pretrained Foundation Models like BERT~\cite{devlin2019bert} and GPT, which ``\textit{are regarded as the foundation for various downstream tasks}''~\cite{zhou2023comprehensive}.  BERT has experienced a surge in popularity over the last few years, and the family of BERT models has repeatedly provided state-of-the-art (SOTA) results for computational modeling of semantic change~\cite{cassotti2023xl,periti2023studying}. However, research focus is now shifting toward ChatGPT due to its impressive ability to generate fluent and high-quality responses to human queries, making it the fastest-growing AI tool. Several recent research studies have assessed the language capabilities of ChatGPT by using a wide range of prompts to solve popular NLP tasks~\cite{laskar2023systematic,KOCON2023ChatGPT}. However, current evaluations generally (a) overlook the fact that the output of ChatGPT is nondeterministic,\footnote{\url{platform.openai.com/docs/guides/gpt/faq}} (b) rely only on contemporary and synchronic text, and (c) consider predictions generated by the ChatGPT\footnote{\url{chat.openai.com}} web interface, which is based on the Chat version of the GPT foundation model. As a result, these evaluations provide valuable insights into the generative, pragmatic, and semantic capabilities of ChatGPT~\cite{KOCON2023ChatGPT}, but fall short when it comes to assess the potential of GPT to solve NLP tasks and specifically to handle historical and diachronic text, which constitutes a unique scenario for testing models' ability to generalize. 

%Nina: I added this but I'm a bit unsure about it. 
%Francesco: We are using to much space for the intro. I don't think there is a need to emphasize more
%In this paper, we evaluate ChatGPT in comparison to BERT in this unique scenario and test the models' ability to generalize to historical text and a very challenging research task. Because of the costs associated with using a model like ChatGPT, we believe that our work can help other researchers evaluate the model for their semantic change task. 

%@Francesco, here would be a good place to put in some aspects of: Our experiments show that ... and then some example on tips and tricks on how to use GPT (and to be a bit cautious because it is not hte same as CHatGPT (with hisotry of prompts etc). 

%\setlength{\belowcaptionskip}{-15pt}
\begin{figure}[!t]
\centering
\includegraphics[width=\columnwidth]{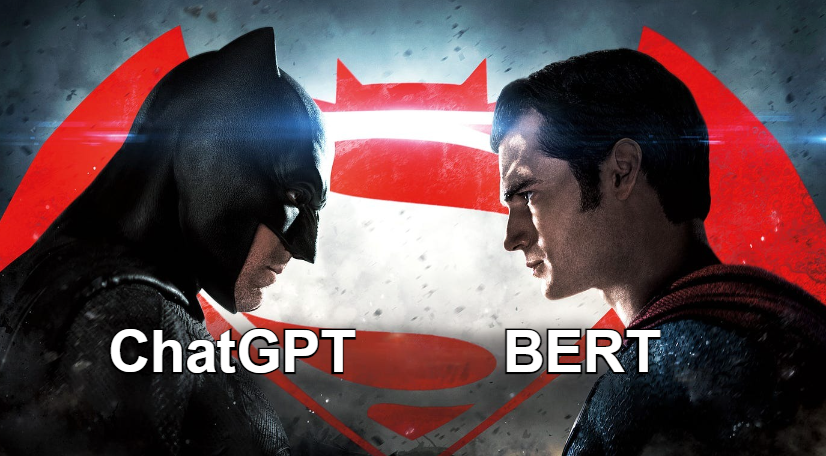}
\caption{The title of this paper draws inspiration by the movie \textit{Batman v Superman: Dawn of Justice}. We leverage the analogy of (Chat)GPT and BERT, powerful and popular LMs, as two lexical superheroes often erroneously associated for solving similar problems. Our aim is to shed lights on the potential of (Chat)GPT for semantic change detection.}
\label{fig:meme}
\end{figure}

In this paper, we propose to evaluate the use of both ChatGPT and GPT - i.e., (Chat)GPT\footnote{Throughout the text, we distinguish between ChatGPT, which is the standard (web) version of GPT, and GPT, which serves as the foundation model. Instances of (Chat)GPT represent both types of models.} - to recognize (lexical) semantic change. Our goal is not to comprehensively evaluate (Chat)GPT in dealing with semantic change but rather to evaluate its potential as \textit{off-the-shelf} model with a \textit{reasonable} prompts from a human point of view, which may not necessarily be optimized for the model. Recently, a novel evaluation task in NLP, called Lexical Semantic Change (LSC), has been introduced as a shared task at SemEval~\cite{schlechtweg2020semeval}. The LSC task involves considering all occurrences (potentially several thousands) of a set of target words to assess their change in meaning within a diachronic corpus. As a result, this setup is \textit{currently} not suitable for evaluating a GPT model, due to the limited size of its prompts and answers, as well as accessibility limitations such as an hourly character limit and economic constraints. In light of these considerations, we chose to evaluate the potential of (Chat)GPT  through the Word-in-Context (WiC,~\citealp{pilehvar2019wic}) task, which has recently demonstrated a robust connection with LSC~\cite{cassotti2023xl,arefyev2021deep}. In particular, we consider two diachronic extensions of the original WiC setting, namely \textit{temporal} WiC (TempoWiC,~\citealp{loureiro2022tempowic}) and \textit{historical} WiC (HistoWiC). Our goal is to determine whether a word carries the same meaning in two different contexts of different time periods, or conversely, whether those contexts exemplify a semantic change. While TempoWiC has been designed to evaluate LMs ability to detect short-term changes in social media, HistoWiC is our adaptation of the SemEval benchmark of historical text to a WiC task for evaluating LMs ability to detect long-term changes in historical corpora. 

Considering the remarkable performance of contextualized BERT models in addressing WiC and LSC tasks~\cite{montanelli2023survey,periti2023time,periti2023studying}, we compare the performance of (Chat)GPT in TempoWiC and HistoWiC to those obtained using BERT. While BERT is specifically designed to understand the meaning of words in context, (Chat)GPT is designed to generate fluent and coherent text. Through these two \textit{lexical superheros} (see Figure~\ref{fig:meme}), we aim to illuminate the potential of (Chat)GPT as \textit{off-the-shelf} model and mark \textit{the dawn of a new era} by  assessing whether it \textit{already} makes the approaches to WiC and LSC, which rely on BERT-embedding similarities, outdated.

\section{Related work}
The significant attention garnered by ChatGPT has led to a large number of studies being published immediately after its release. Early studies mainly focused on exploring the benefits and risks associated with using ChatGPT in expert fields such as education~\cite{lund2023chatting}, medicine~\cite{ANTAKI2023100324}, or business~\cite{Shaji2023Review}. Evaluation studies are currently emerging for assessing (Chat)GPT's generative and linguistic  capabilities across a wide range of downstream tasks in both monolingual and multilingual setups~\cite{bang2023multitask,shen2023hugginggpt,lai2023chatgpt}. Most evaluations focus on ChatGPT and involve a limited number of instances (e.g., 50) for each task considered~\cite{weissweiler2023counting,zhong2023chatgpt,alberts2023large,khalil2023chatgpt}. When the official API is used to query the GPT foundation model, this limit is imposed by the hourly token processing limit\footnote{\url{help.openai.com/en/articles/4936856-what-}\\\url{are-tokens-and-how-to-count-them}} and the associated costs.\footnote{\url{openai.com/pricing}} When the web interface is used instead of the API, the limit is due to the time-consuming process of interacting with ChatGPT that keeps humans in the loop. Thus far, even systematic and comprehensive evaluations ~\cite{KOCON2023ChatGPT,laskar2023systematic} rely on repetition of a single experiment for each task. However, while individual experiments provide valuable insights into (Chat)GPT's capabilities, they fall short in assessing the potential of (Chat)GPT to solve specific tasks given its nondeterministic nature. Multiple experiments need to be conducted to validate its performance on each task. In addition, current  evaluations generally leverage tasks that overlook the temporal dimension of text, leaving a gap in our understanding of (Chat)GPT's ability to handle diachronic and historical text.\\

\noindent \textbf{Our original contribution.} \\ Our evaluation of (Chat)GPT focuses on two diachronic extensions of the WiC task, namely TempoWiC and HistoWiC. Our aim is to assess the potential of (Chat)GPT for \textbf{Semantic Change Detection}. To the best of our knowledge, this paper is the first to investigate the application of (Chat)GPT for historical linguistic purposes. Thus far, only the use of ChatGPT for a conventional WiC task has been evaluated by~\newcite{laskar2023systematic} and~\newcite{KOCON2023ChatGPT}, who reported low accuracy % of 62.1\% and 64.58\%, 
under a single setup. In this paper, we challenge their performance by considering diachronic text and the following setups, totaling 47 experiments each for TempoWiC and HistoWiC:
\begin{itemize}
    \item \textbf{Different prompts}. \ Like~\newcite{zhong2023chatgpt}, we evaluate (Chat)GPT using zero-shot and few-shot prompting strategies, while also exploring many-shot prompting. Our results demonstrate that zero-shot prompting is more effective on HistoWiC, while few-shot prompting is more effective on TempoWiC. 

    \item \textbf{Varying temperature}. \ Like~\newcite{peng2023making,liu2023code}, we analyze how GPT's performance varies according to its temperature hyperparameter, which controls the ``creativity'' or randomness of its answers. Our results indicate that GPT used with low temperature values (i.e., less creativity) is better at handling WiC tasks. 
    
    \item \textbf{GPT API v ChatGPT Web}. \ We empirically assess whether GPT produces worse results through the OpenAI API compared to ChatGPT through the web interface.\footnote{Discussions on this topic are currently very active, for example,~\url{community.openai.com/t/web-app-vs-api-}\\\url{results-web-app-is-great-api-is-pretty-awful/96238}} Our results demonstrate that using GPT through the official API for WiC tasks is better than using ChatGPT through the web interface, as has previously been done~\cite{laskar2023systematic,KOCON2023ChatGPT}. Furthermore, our findings suggest that the web interface automatically sets an intermediate temperature for ChatGPT.

    \item \textbf{(Chat)GPT v BERT}. \ Finally, like~\newcite{zhong2023chatgpt}, we compare the performances of (Chat)GPT and BERT. By leveraging the TempoWiC task and introducing the novel HistoWiC task, we shed light on the potential of both models and demonstrate the \textit{current} superiority of BERT in dealing with diachronic text and WiC tasks, compared to \textit{reasonable} GPT prompts templates and strategies.
\end{itemize}

\section{Semantic Change Detection}
Our evaluation relies on two diachronic definitions of the conventional Word-in-Context (WiC) task, namely TempoWiC and HistoWiC. WiC is framed as a binary classification problem, where each instance is associated with a target word $w$,  either a verb or a noun, for which two contexts, $c_1$ and $c_2$, are provided. The task is to identify whether the occurrences of $w$ in $c_1$ and $c_2$ correspond to the same meaning or not. Both TempoWiC and HistoWiC rely on the same definition of the task, while being specifically designed for semantic change detection in diachronic text.

\subsection{Temporal Word-in-Context}
NLP models struggle to cope with new content and trends. TempoWiC is designed as an evaluation benchmark to detect short-term semantic changes on social media, where the language is extremely dynamic.  It  uses tweets from different time periods as contexts $c_1$ and $c_2$.

Given the limits on testing (Chat)GPT, we followed~\newcite{zhong2023chatgpt,jiao2023chatgpt} and randomly sampled a subset of the original TempoWiC datasets. While the original TempoWiC framework provides trial, train, test, and dev sets, here we did not consider the dev set. Table~\ref{tab:datasets} shows the number of positive (i.e., same meaning) and negative (i.e., different meanings due to semantic change) examples we considered for each set.

\begin{table}[!ht]
\centering
\caption{Datasets used in our evaluation} 
\resizebox{0.8\columnwidth}{!}{
\begin{tabular}{ccccccc}
                                   & \multicolumn{3}{c}{\textbf{TempoWiC}}     & \multicolumn{3}{c}{\textbf{HistoWiC}}     \\ \cline{2-7} 
\multicolumn{1}{c|}{}              & Trial & Train & \multicolumn{1}{c|}{Test} & Trial & Train & \multicolumn{1}{c|}{Test} \\ \cline{2-7} 
\multicolumn{1}{c|}{\textit{True}} & 8     & 86    & \multicolumn{1}{c|}{73}   & 11    & 137   & \multicolumn{1}{c|}{79}   \\
\multicolumn{1}{c|}{\textit{False}} & 12 & 114 & \multicolumn{1}{c|}{127} & \textit{9} & 103 & \multicolumn{1}{c|}{61} \\ \cline{2-7} 
\multicolumn{1}{c|}{\textbf{Total}}  & 20    & 200   & \multicolumn{1}{c|}{200}  & 20    & 200   & \multicolumn{1}{c|}{140}  \\ \cline{2-7} 
\end{tabular}}
\label{tab:datasets}
\end{table}

\subsection{Historical Word-in-Context}
Given that NLP models also struggle to cope with historical content and trends, we designed HistoWiC as a novel evaluation benchmark for detecting long-term semantic change in historical text, where language may vary across different epochs. HistoWiC sets the two contexts, $c_1$ and $c_2$, as sentences collected from the two English corpora of the LSC detection task ~\cite{schlechtweg2020semeval}. 

Similar to the original WiC~\cite{pilehvar2019wic}, the annotation process for the LSC benchmark involved usage pair annotations where a target word is used in two different contexts. Thus, we directly used the annotated instances of LSC to develop HistoWiC. Since LSC instances were annotated  using the DURel framework~\cite{schlechtweg2023durel} and a four-point semantic-relatedness scale~\cite{schlechtweg2021dwug,schlechtweg2020semeval,schlechtweg2018diachronic}, we only binarized the human annotations (see Appendix~\ref{app:fromLSCtoHistoWiC}).

As with TempoWiC, we randomly sampled a limited number of instances to create trial, training, and test sets. Table~\ref{tab:datasets} shows the number of positive and negative examples for each set.

\section{Experimental setup}
In the following, we present our research questions (RQs) and the various setups we considered in our work. In our experiments, we evaluated the performance of (Chat)GPT 3.5 over the TempoWiC and HistoWiC test sets using both the official OpenAI API (GTP API)\footnote{version 0.27.8.} and the web interface (ChatGPT Web).\footnote{The August 3 Version.} Of the GPT 3.5 models available through the API, we assessed the performance of \verb|gpt-3.5-turbo|. Following~\newcite{loureiro2022tempowic}, we employed the Macro-F1 for multiclass classification problems as evaluation metric. 

\subsection{(Chat)GPT prompts} 
Current ChatGPT evaluations are typically performed manually~\cite{laskar2023systematic}. When automatic evaluations are performed, they are typically followed by a manual post-processing procedure~\cite{KOCON2023ChatGPT}. As manual evaluation and processing may be biased due to answer interpretation, we addressed the following research question:

\paragraph{RQ1:} \textit{Can we evaluate (Chat)GPT in WiC tasks  in a completely automatic way?}\\

Furthermore, as current evaluations generally rely on a zero-shot prompting strategy, we addressed the following research question:

\paragraph{RQ2:} \textit{Can we enhance (Chat)GPT's performance in WiC tasks by leveraging its in-context learning capabilities?} \\

To address RQ1 and RQ2, we designed a prompt template to explicitly instruct (Chat)GPT to answer in accordance with the WiC label format (i.e., \textit{True}, \textit{False}). We then used this template (see Appendix~\ref{app:WiCtemplate}) with different prompt strategies: 
\begin{compactitem}
    \item \textit{zero-shot prompting} (ZSp): (Chat)GPT was asked to address the WiC tasks (i.e., test sets) without any specific training, generating coherent responses based solely on its preexisting knowledge.
    \item \textit{few-shot prompting} (FSp): since PFMs have recently demonstrated \textit{in-context learning} capabilities without requiring any fine-tuning on task-specific data~\cite{Brown2020Language}, (Chat)GPT was presented with a limited number of input-output examples (i.e., trial sets) demonstrating how to perform the task. The goal was to leverage the provided examples to improve the model's task-specific performance.
    \item \textit{many-shot prompting} (MSp): similar to FSp, but with a greater number of input-output examples (i.e., training sets).
\end{compactitem}

\subsection{(Chat)GPT temperature}
The temperature is a hyperparameter of (Chat)GPT that regulates the variability of responses to human queries. According to the OpenAI FAQ, the temperature parameter ranges from 0.0 to 2.0, with lower values making outputs mostly deterministic and higher values making them more random.\footnote{\url{platform.openai.com/docs/api-reference/chat}} To counteract the nondeterminism of (Chat)GPT, we focused only on TempoWiC and HistoWiC and conducted the same experiment multiple times with progressively increasing temperatures. This approach enabled us to answer the following research questions:

\paragraph{RQ3:} \textit{Does (Chat)GPT demonstrate comparable effectiveness in detecting short-term changes in contemporary text and long-term changes in historical text?}\\

\paragraph{RQ4:} \textit{Can we enhance (Chat)GPT's performance in WiC tasks by raising the ``creativity'' using the temperature value?}\\

To address RQ3 and RQ4, we evaluated GPT API in TempoWiC and HistoWiC using eleven temperatures in the range [0.0, 2.0] with 0.2 increments. %\footnote{$\{0.0, 0.2, 0.4, 0.6, 0.8, 1.0, 1.2, 1.4, 1.6, 1.8, 2.0\}$} 
For each temperature and prompting strategy, we performed two experiments and considered the average performance. 

\subsection{GPT API v ChatGPT Web}
Current evaluations typically prompt GPT through the web interface instead of the official OpenAI API. This preference exists because the web interface is free and predates the official API. However, there are differences between using ChatGPT through the web interface (ChatGPT Web) and  the official API (GPT API). First of all, the official API enables to query the GPT foundation  model, while the web interface enables to query the Chat version. In addition, the GPT API can be set to test at varying temperatures, but the temperature value on ChatGPT Web cannot be controlled. However,  while the GPT API allows a limited message history, ChatGPT Web seems to handle an unlimited message history (see Appendix~\ref{app:history}).

We used the following research question to compare the performance of GPT API and ChatGPT Web:

\paragraph{RQ5:} \textit{Does GPT API demonstrate comparable performance to ChatGPT Web in solving WiC tasks?}\\

Testing GPT API with the MSp strategy would be equivalent to testing it with the FSp strategy due to the limited message history. Thus, we evaluated ChatGPT Web with MSp,  aiming to address the following research question:

\paragraph{RQ6:} \textit{Can we enhance ChatGPT's performance in WiC tasks by providing it with a larger number of in-context examples?}\\

To address these research questions, we tested (Chat)GPT using a single chat for each prompting strategy considered. Since testing ChatGPT Web is extremely time-consuming, we conducted one experiment for each prompting strategy. 

\subsection{(Chat)GPT v BERT}
The ability of (Chat)GPT to understand has prompted the belief that ChatGPT is a \textit{jack of all trades} that makes previous technologies somewhat outdated. Drawing upon~\newcite{KOCON2023ChatGPT}, we believe that, when used for solving downstream tasks as \textit{off-the-shelf} model,  (Chat)GPT is \textit{currently} a \textit{master of none.} It works on a comparable level to the competition, but does not outperform any major SOTA solutions.

By relying on multiple experiments on TempoWiC and HistoWiC, we aimed to empirically assess the potential of (Chat)GPT for WiC and LSC tasks. In particular, we addressed the following research question:

\paragraph{RQ7:} \textit{Does (Chat)GPT outperform BERT embeddings in detecting semantic changes?} \\

To address RQ7, we evaluated \verb|bert-base-uncased| on TempoWiC and HistoWiC over different layers. Recent research has exhibited better results when utilizing earlier layers rather than the final layers for solving downstream tasks such as WiC~\cite{periti2023time,ma2019universal,coenen2019visualizing,liang2023named}. For each layer, we extracted the word embedding for a specific target word $w$ in the context $c_1$ and $c_2$. Since the focus of our evaluation was on (Chat)GPT, we did not fine-tune BERT and simply used the similarity between the embeddings of $w$ in the context $c_1$ and $c_2$. In particular, we followed~\newcite{pilehvar2019wic}, and trained a threshold-based classifier using the cosine distance between the two embeddings of each pair in the training set. The training process consisted of selecting the threshold that maximized the performance on the training set. We trained a distinct threshold-based classifier for each BERT layer and for each WiC task (i.e., TempoWiC and HistoWiC). Then, in our evaluation, we applied these classifiers to evaluate BERT over the TempoWiC and HistoWiC test sets.

Finally, we addressed the following research question:

\paragraph{RQ8:} \textit{Can we rely on the pretrained knowledge of GPT to automatically solve the LSC task?}\\

Since (Chat)GPT has demonstrated awareness of historical lexical semantic changes when manually asked about the lexical semantic changes of some words (e.g., \textit{plane}), our goal with RQ8 was to automatically test GPT's pretrained knowledge of historical semantic changes covered in the English LSC benchmark. In addressing this research question we relied on the LSC ranking task as defined in~\newcite{schlechtweg2018diachronic}. Thus, we specifically asked GPT to rank the set of 37 target words in the English LSC benchmark according to their degree of LSC between two time periods, T1 (1810--1860) and T2 (1960--2010). For each temperature, we repeated the same experiment ten times, totaling 110 experiments. Then, for each temperature, we evaluated GPT's performance by computing the Spearman correlation using gold scores derived from human annotation and the average GPT score for each target (see Appendix~\ref{app:LSCtemplate}). 

\section{Experimental results}
In this section, we report the results of our experiments, while discussing the findings in regard to each research question.\footnote{We provide all our data, code, and results at \url{https://github.com/FrancescoPeriti/ChatGPTvBERT}}

\paragraph{RQ1:} %CAMERA READY\textit{Can we evaluate ChatGPT in WiC tasks  in a completely automatic way?}\\

\begin{figure}[!t]
\centering
\includegraphics[width=0.8\columnwidth]{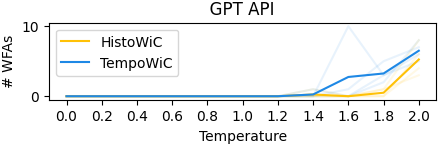}
\caption{Average number of wrongly formatted answers (WFAs) over the  temperature values considered. Background lines correspond to each experiment.}
\label{fig:wfa}
\end{figure}

(Chat)GPT consistently followed our template in nearly all cases, thereby allowing us to evaluate its answers without human intervention. For GPT API, however, we noticed that the higher the temperature, the larger the tendency for deviations from the expected response format (see Figure~\ref{fig:wfa}). ChatGPT Web only once answered with an incorrect format. To ensure impartiality, we classified the few (Chat)GPT responses that did not adhere to the required format as incorrect answers. %%CAMERA READY: \\

\begin{figure}[!t]
\centering
\includegraphics[width=0.7\columnwidth]{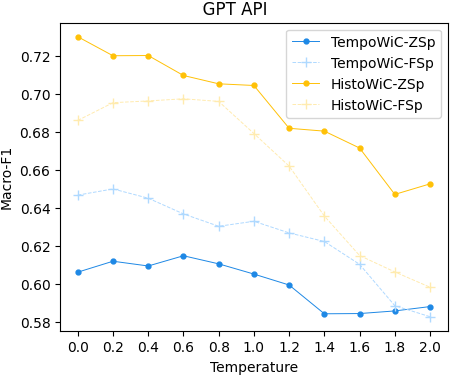}
\caption{Performance of GPT API (Macro-F1) as temperature increases.} 
%\caption{Performance of ChatGPT API (Macro-F1) for TempoWiC and HistoWiC tasks with varying temperature values and prompting strategies}.
\label{fig:api-temperature}
\end{figure}

\paragraph{RQ2:} %%CAMERA READY: \textit{Can we enhance ChatGPT's performance in WiC tasks by leveraging its in-context learning capabilities?} \\

\begin{figure}[!t]
\centering
\includegraphics[width=0.7\columnwidth]{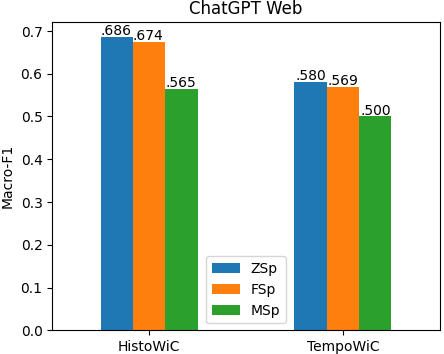}
\caption{Performance of ChatGPT Web (Macro-F1). Temperature is unknown.} %\caption{Performance of ChatGPT Web  (Macro-F1) for TempoWiC and HistoWiC tasks with varying prompting strategies. Temperature is unknown for ChatGPT Web.}
\label{fig:web-prompt}
\end{figure}

Figure~\ref{fig:api-temperature} shows the rolling average of the performance  of GPT API across different temperatures, prompting strategies, and WiC tasks. By using a window size of 4, we were able to consider 8 different experiments per temperature (for each temperature, we ran two experiments)\footnote{Except for the first and last two temperatures.}. Figure~\ref{fig:web-prompt} shows the performance of ChatGPT Web across different prompting strategies and WiC tasks.

Figure~\ref{fig:api-temperature} and~\ref{fig:web-prompt} show that ZSp consistently outperforms FSp on HistoWiC. By contrast, FSp consistently outperforms ZSp in TempoWiC when the GPT API is used. This result suggests that the in-context learning capability of GPT is more limited for historical data. In Figure~\ref{fig:web-prompt}, ChatGPT Web's performance with ZSp outperforms that obtained with FSp for both TempoWiC and HistoWiC, although the discrepancy is smaller. 

\paragraph{RQ3:} %%CAMERA READY: \textit{Does ChatGPT demonstrate comparable effectiveness in detecting short-term changes in contemporary text and long-term changes in historical text?}\\

\begin{table}[!ht]
\centering
\caption{Macro-F1 scores obtained by SOTA systems, (Chat)GPT (best score), and BERT (last layer).}
%\caption{Comparison of (Macro-F1) scores obtained by SOTA systems on the full TempoWiC dataset and scores %obtained on our TempoWiC subset using the ChatGPT API, ChatGPT Web, and BERT. For ChatGPT API and Web, we report %the best score obtained in our experiments. For BERT, we report the score obtained using the last layer.}
\resizebox{0.5\columnwidth}{!}{%
\begin{tabular}{cc}
 & \textbf{Macro-F1} \\ \hline
\citealp{chen2022using} & .770 \\
\citealp{loureiro2022tempowic} & .703 \\
\citealp{loureiro2022tempowic} & .670 \\
\citealp{lyu2022mllabs} & .625 \\
\textit{GPT API} & .689 \\
\textit{ChatGPT Web} & .580 \\
\textit{BERT} & .743 \\
\end{tabular}%
}
\label{tab:sota-tempowic}
\end{table}

Figures~\ref{fig:api-temperature} and~\ref{fig:web-prompt} show that (Chat)GPT's performance on TempoWiC is consistently lower than its performance on HistoWiC. In particular, in our experiments we observe that (Chat)GPT's performance ranges from .551 to .689 on TempoWiC and from .552 to .765 on HistoWiC. This suggests that (Chat)GPT is significantly more effective for long-term change detection than for short-term change detection. We believe that this might involve word meanings that were not explicitly covered during training, potentially allowing (Chat)GPT to detect anomalies from the usual patterns. We will further investigate this aspect in our future research.

For the sake of comparison, we report SOTA performance in Table~\ref{tab:sota-tempowic}. Results from this research are in italics.

\paragraph{RQ4:} %%CAMERA READY\textit{Can we enhance ChatGPT's performance in WiC tasks by raising the ``creativity'' using the temperature value?}\\

Figure~\ref{fig:api-temperature} shows that, on average, higher performance is associated with lower temperatures for both TempoWiC and HistoWiC, with accuracy decreasing as temperature values increase. Thus, we argue that high temperatures do not make it easier for GPT to solve WiC tasks or identify semantic changes effectively.

\paragraph{RQ5:} %%CAMERA READY\textit{Does ChatGPT API demonstrate comparable performance to ChatGPT Web in solving WiC tasks?}\\

ChatGPT Web results are presented in Table~\ref{tab:web}, along with the average performance we obtained through the GPT API across temperature values ranging from 0.0 to 1.0 (API 0--1), from 1.0 to 2.0 (API 1--2), and from 0.0 to 2.0 (API 0--2). As with GPT API, the performance of ChatGPT Web is higher for HistoWiC than for TempoWiC. In addition, our evaluation indicates that ChatGPT Web employs a moderate temperature setting, for we obtained consistent results when using a moderate temperature setting through GPT API. This suggests that the GPT API should be preferred  for solving downstream task like WiC. It also suggests that the current SOTA evaluations may achieve higher results if the official API were used instead of the web interface. Thus, this implies that previous results using web interface should be interpreted with caution.

\begin{table}[!ht]
\centering
\caption{Comparison of GPT API and ChatGPT Web performance (Macro-F1)} %\caption{Comparison of ChatGPT Performance (Macro-F1) for TempoWiC and HistoWiC tasks using ChatGPT API and ChatGPT Web. We consider the average scores obtained over various temperatures as ChatGPT API values.}
\resizebox{0.8\columnwidth}{!}{%
\begin{tabular}{ccccccccc}
 & \multicolumn{4}{c}{\textbf{TempoWiC}} & \multicolumn{4}{c}{\textbf{HistoWiC}} \\ \cline{2-9} 
\multicolumn{1}{c|}{} & \textit{API} & \textit{API} & \textit{API} & \multicolumn{1}{c|}{\textit{web}} & \textit{API} & \textit{API} & \textit{API} & \multicolumn{1}{c|}{\textit{web}} \\
\multicolumn{1}{c|}{Temp.} & 0--1 & 1--2 & 0--2 & \multicolumn{1}{c|}{-} & 0--1 & 1--2 & 0--2 & \multicolumn{1}{c|}{-} \\ \hline
\multicolumn{1}{c|}{ZSp} & .609 & .589 & .600 & \multicolumn{1}{c|}{.580} & .713 & .665 & .688 & \multicolumn{1}{c|}{.686} \\
\multicolumn{1}{c|}{FSp} & .636 & .606 & .622 & \multicolumn{1}{c|}{.569} & .693 & .626 & .657 & \multicolumn{1}{c|}{.674} \\
\multicolumn{1}{c|}{MSp} & - & - & - & \multicolumn{1}{c|}{.500} & - & - & - & \multicolumn{1}{c|}{.565} \\ \hline
\multicolumn{1}{c|}{\textbf{all}} & .622 & .598 & .611 & \multicolumn{1}{c|}{.550} & .703 & .645 & .672 & \multicolumn{1}{c|}{.642}\\ \cline{2-9} 
\end{tabular}%
}
\label{tab:web}
\end{table}

\paragraph{RQ6:} %%CAMERA READY\textit{Can we enhance ChatGPT's performance in WiC tasks by providing it with a larger number of in-context examples?}\\

As shown in Figure~\ref{fig:web-prompt},  the performance of (Chat)GPT Web decreases as the number of example messages increases (from ZSp to MSp). This suggests that improving the performance of (Chat)GPT requires a more complex training approach than simply providing a few input-output examples. Furthermore, it indicates that the influence of message history is extremely significant in shaping the quality of conversations with (Chat)GPT. Indeed, a limited message history proved to be beneficial for the evaluation of GPT API through FSp.

\paragraph{RQ7:} %%CAMERA READY\textit{Does ChatGPT outperform BERT embeddings in detecting semantic changes?} \\

Figure~\ref{fig:bert-layers} shows Macro-F1 scores obtained on TempoWiC and HistoWiC over the 12 BERT layers (see Appendix~\ref{app:bert}). When considering the final layer, which is conventionally used in downstream tasks, BERT obtains Macro-F1 scores of .750 and .743 for TempoWiC and HistoWiC, respectively. Similar to~\newcite{periti2023time}, BERT performs best on HistoWiC when embeddings extracted from middle layers are considered. However, BERT performs best on TempoWiC when embeddings extracted from the last layers are used. 

\begin{figure}[!t]
\centering
\includegraphics[width=0.7\columnwidth]{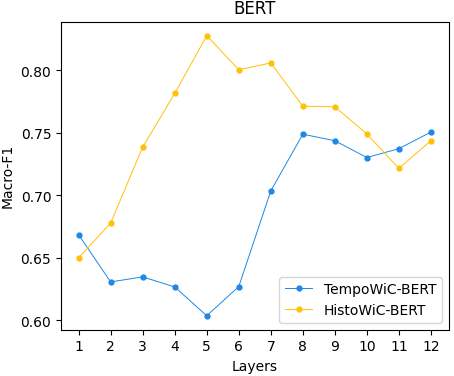}
\caption{Comparison of BERT Performance (Macro-F1) for TempoWiC and HistoWiC tasks across layers}
\label{fig:bert-layers}
\end{figure}

\begin{figure}[!t]
\centering
\includegraphics[width=0.6\columnwidth]{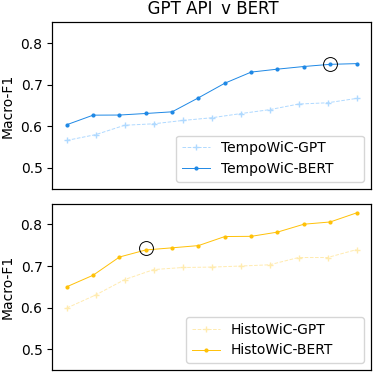}
\caption{GPT v BERT (Macro-F1). Performance is sorted in ascending order regardless of temperatures and layers. A black circle denotes the use of the last layer of BERT.}%\caption{Comparison of BERT and ChatGPT performance (Macro-F1). Performance is sorted in ascending order regardless of temperatures and layers. For comparison, we marked  the performance of BERT when the last layer is used for embedding extraction with a black circle.}
\label{fig:gptvbert}
\end{figure}

We compared the performance of GPT and BERT across their respective worst to best scenarios by sorting the Macro-F1 scores obtained by BERT and GPT in ascending order (bottom x-axis). For ChatGPT, we consider the results obtained through FSp and ZSp prompting for TempoWiC and HistoWiC, respectively. As shown in Figure~\ref{fig:gptvbert}, even when considering the best setting, GPT does not outperform the Macro-F1 score obtained by using the last layer of BERT, marked with a black circle. However, although it exhibits lower performance, the results obtained from GPT are still comparable to BERT results on HistoWiC when embeddings extracted from the last layer of BERT are used.

\begin{figure}[!t]
\centering
\includegraphics[width=0.8\columnwidth]{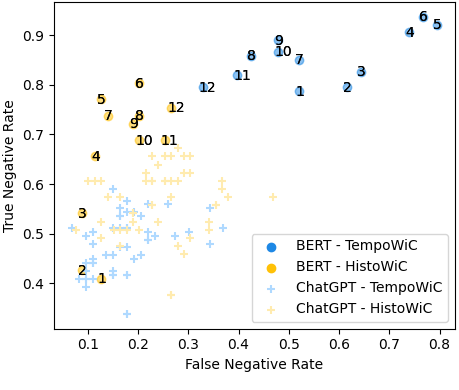}
\caption{True Negative Rate v False Negative Rate. Each cross represents a (Chat)GPT experiment. Each dot represents the use of a specific layer of BERT.}%\caption{True Negative Rate v False Negative Rate using ChatGPT and BERT. Each cross represents a ChatGPT experiment on TempoWiC and HistoWiC. Each dot represents the use of a specific layer of BERT on TempoWiC and HistoWiC}
\label{fig:tnrfnr}
\end{figure}

Since our goal is to evaluate the potential of (Chat)GPT for recognizing lexical semantic changes, we analyzed the true negative rate and false negative rate scores, because \textit{negative} examples represent semantic change in TempoWiC and HistoWiC datasets. As shown in Figure~\ref{fig:tnrfnr}, regardless of the temperature and layer considered, (Chat)GPT falls short in recognizing semantic change for both TempoWiC and HistoWiC compared to BERT. However, it produces fewer false negatives than BERT for TempoWiC.\\

\paragraph{RQ8:} %%CAMERA READY\textit{Can we rely on the pretrained knowledge of ChatGPT to automatically solve the LSC task?}\\

In our experiment, GPT achieved low Spearman’s correlation coefficients for each temperature when ranking the target word of the LSC English benchmark by degree of lexical semantic change. Higher correlations were achieved by using low temperatures rather than high ones (see Appendix~\ref{app:lsc}).
Table~\ref{tab:LSC} shows the GPT correlation for the temperature 0. For comparison, we report correlations obtained by BERT-based systems that leverage pretrained models. Note that, when BERT is fine-tuned, it generally achieves even higher correlation scores (see survey by~\citealp{montanelli2023survey}).

\begin{table}[!ht]
\centering
\caption{LSC comparison: correlation obtained by SOTA, \textit{pre-trained} BERT systems and GPT (temperature=0).}
%\caption{Comparison of correlation obtained by SOTA BERT-based systems and correlation obtained by ChatGPT with temperature set to 0.}
\resizebox{0.75\columnwidth}{!}{%
\begin{tabular}{cc}
 & \textbf{Spearman’s correlation} \\ \hline
\citealp{periti2023studying} & .651 \\ 
\citealp{laicher2021explaining} & .573 \\
\citealp{periti2022done} & .512 \\
\citealp{rother2020cmce} & .512 \\ 
\textit{GPT API} & .251
\end{tabular}%
}
\label{tab:LSC}
\end{table}

As shown in Table~\ref{tab:LSC}, the BERT-based system largely outperforms GPT, suggesting that GPT is not currently well-adapted for use in solving LSC downstream tasks.

\subsection{BERT for Semantic Change Detection}
There are notable differences between the Macro-F1 for TempoWiC and HistoWiC in terms of how the results increase and decrease across layers (see Figure~\ref{fig:bert-layers}). For TempoWiC the results increase until the 8th layer, after which they remain almost stable. Conversely, for HistoWiC the BERT performance rapidly increases until the 5th layer, after which it linearly decreases until the 12th layer. As regards Tempo WiC, we hypothesize that BERT is already aware of the set of word meanings considered for evaluation as it was pretrained on modern and contemporary texts. As regards HistoWiC, we hypothesize that BERT is not completely aware of the set of word meanings considered for evaluation and that word representations adopted for the historical context of HistoWiC\footnote{1810--1860, as referenced in~\newcite{schlechtweg2020semeval}} might be slightly tuned. Thus, using medium embedding layers could prove beneficial in detecting semantic changes, as these layers are less affected by contextualization~\cite{ethayarajh2019contextual}. In other words, for HistoWiC, we hypothesize that the performance diminishes in the later layers due to the increasing contextualization of the medium and final embedding layers, which %hinders 
reduces the presence of noise in untuned word representations. This prompts us to question the appropriateness of using the last embedding layers to recognize historical lexical semantic change. We will address this question in  future research.

\section{Conclusion}
In this study, we empirically investigated the use of the \textit{current} (Chat)GPT 3.5 to detect semantic change. Our goal is not to comprehensively evaluate (Chat)GPT in dealing with semantic change, but rather to acknowledge its potential while also raising concerns and questions about its off-the-shelf use. In this regard, we used \textit{reasonable} prompts from a human point of view, which may not necessarily be optimized for the model. We used the TempoWiC benchmark to assess (Chat)GPT's ability to detect short-term semantic changes, and introduced a novel benchmark, HistoWiC, to assess (Chat)GPT's ability to recognize long-term changes. When considering the standard 12 layer of BERT, our experiments show that (Chat)GPT achieves comparable performance to BERT (although slightly lower) in regard to detecting long-term changes, but performs significantly worse in regard to recognizing short-term changes. We find that BERT's contextualized embeddings consistently provide a more effective and robust solution for capturing both short- and long-term changes in word meanings.

There are two possible explanations for the discrepancy in (Chat)GPT's performance between TempoWiC and HistoWiC: i) HistoWiC might involve word meanings not explicitly covered during training, potentially aiding (Chat)GPT in detecting anomalies; ii) TempoWiC involves patterns typical of Twitter (now X), such as abbreviations, mentions, or tags, which may render it more challenging than HistoWiC.

In light of our findings, we argue that \textit{(Chat)GPT 3.5 might be the hero the world deserves but not the one it needs right now}\footnote{This quote draws inspiration by
the movie \textit{Batman: The Dark Knight}. We leverage the analogy of (Chat)GPT achieving lower results than BERT to acknowledge the potential of (Chat)GPT while also raising concerns and questions about its use for Semantic Change detection.}, in particular for computationally modeling meaning over time, and by extension, for the study of semantic change. Nevertheless, during the course of our research, updates to (Chat)GPT became available and gained popularity, leading research and practitioners to conduct new experiments on these updated models. Particularly noteworthy is a recent study by~\newcite{karjus2023machineassisted}, which showcased remarkable performance on LSC using the GPT-4 model. Inspired by this research, our ongoing and future work is focused on further exploring the capabilities of GPT-4 for modeling semantic change.

\section*{Limitations}
There are limitations we had to consider in the making of this paper. Firstly, a limitation arises when working with temporal HistoWiC benchmarks. While we ensure the utilization of diachronic data, we cannot guarantee that if the meaning of a word differs across contexts, it unequivocally indicates either the presence of stable polysemy (existing stable multiple meanings) or exemplifies a semantic change (either a new sense that it did not previously possess or a lost sense that it no longer has).

Other limitations are about the use of language models. We could not evaluate (Chat)GPT across different languages due to both price and API limitations. This means that while the results holds for English, we do not know how (Chat)GPT will behave for the other languages. Although we are aware of open source solution such as LLaMA, it still necessitates expensive research infrastructure, and we thus chose to focus on (Chat)GPT. 

Like all research on (Chat)GPT~\cite{laskar2023systematic, KOCON2023ChatGPT, zhong2023chatgpt}, our work has a significant limitation that we cannot address: our (Chat)GPT results are not entirely reproducible as (Chat)GPT is inherently nondeterministic. In addition, like ~\newcite{zhong2023chatgpt,jiao2023chatgpt}, we found that time and economic constraints when using (Chat)GPT dictated that our evaluation of the software had to be based on only  a subset of the TempoWiC and HistoWiC dataset. %Furthermore, due to time constraints, we conducted only a single experiment for each prompting strategy when considering ChatGPT by using the web interface. A more convincing approach would involve testing ChatGPT on a larger sample size and conducting multiple rounds of experiments for each temperature and prompt strategy. In future work, we aspire to not only evaluate ChatGPT on a broader range of examples but also to conduct more in-depth analysis and discussion across similar outputs at different temperatures.

In our study, we utilized (Chat)GPT 3.5. This could be considered a limitation, given the recent release of GPT 4. However, we opted for (Chat)GPT 3.5 based on the guidance provided in the current OpenAI documentation.\footnote{\url{https://platform.openai.com/docs/guides/gpt/which-model-should-i-use}} Additionally, we argue that (Chat)GPT-3.5 is a cheaper alternative than the current GPT-4 model, making the investigation of (Chat)GPT-3.5 still significant for researchers with limited economic resources. We acknowledge that OpenAI continues to train and release new models, which could potentially affect the reproducibility of our results.

One of the many features of (Chat)GPT is its ability to incorporate the history of preceding messages within a conversation while responding to new input prompts. However, there remain several unanswered questions regarding how this history influences the model's answers. This holds true even for the zero-shot prompting strategy, where a general setting is lacking. Multiple prompts can be provided as part of the same chat or across different chats. For simplicity, and similar to previous research, we assigned only one chat for each ZSp experiment. We intend to use different chats in our future work to examine and investigate the effect of the message history.

Finally, as highlighted by~\newcite{laskar2023systematic}, since the instruction-tuning datasets of OpenAI models are unknown (that is, not open source), the datasets used for evaluation may or may not be part of the instruction-tuning training data of OpenAI. 

Despite these limitations, we argue that our work is significant as it may prompt new discussion on the use of LMs such as BERT and (Chat)GPT, while also dispelling the expanding belief that the use of ChatGPT as \textit{off-the-shelf} model \textit{already} makes BERT an outdated technology.

\section*{Acknowledgements}
This work has in part been funded by the project Towards Computational Lexical Semantic Change Detection supported by the Swedish Research Council (2019–2022; contract 2018-01184), and in part by the research program Change is Key! supported by Riksbankens Jubileumsfond (under reference number M21-0021). The computational resources were provided by the National Academic Infrastructure for Supercomputing in Sweden (NAISS).

\bibliography{bib}

\newpage
\section*{Appendix}
\appendix
\section{Historical WiC}\label{app:fromLSCtoHistoWiC}
We shifted from the LSC to the WiC setting as follows. First, we selected only the annotated LSC instances containing contexts from different time periods. We then filtered out all the instances annotated by a single annotator\footnote{Different instances were annotated by varying numbers of annotators.} and all the instances that are associated with an average score, $s$, such that $1.5 < s < 3.5$, which represents ambiguous cases even for humans. Finally, we binarized the LSC annotations by converting each $s \leq 1.5$ to \textit{False} (i.e. different meanings) and each $s \geq 3.5$ to \textit{True} (i.e. same meaning). We report in Table~\ref{tab:lsc-scale} the four-point semantic-relatedness used to annotate the LSC instances through the DURel framework.

\begin{table}[!ht]
\centering
\resizebox{0.60\columnwidth}{!}{%
\begin{tabular}{lcl}
\multirow{4}{*}{$\Bigg\uparrow$} & 4: & Identical         \\
                            & 3: & Closely related   \\
                            & 2: & Distantly related \\
                            & 1: & Unrelated        
\end{tabular}}
\caption{The DURel relatedness scale used in~\newcite{schlechtweg2020semeval,schlechtweg2018diachronic}}
\label{tab:lsc-scale}
\end{table}

\section{Message history}\label{app:history}
Although one of the many features of (Chat)GPT is its ability to consider the history of preceding messages within a conversation while responding to new input prompts, GPT API and the web version handle message history differently. In GPT API, the message history is limited to a fixed number of tokens (i.e., 4,096 tokens for \verb|gpt-3.5-turbo|); however, we are not aware of how the message history is handled in ChatGPT Web, where an unlimited number of message for chat seems to be supported.

In our experiments, we use a single chat for each considered prompting strategy, both for ChatGPT Web and GPT API. However, in ChatGPT Web, we considered the full message history for the ZSp, FSp, and MSp strategies. Instead, to avoid exceeding the token limit set by the OpenAI API, we tested GPT API for the ZSp and FSp strategies by considering a message history of 33 messages. Note that due to the token limit, testing the MSp strategy for GPT API wasn't possible, as the limited message history would make MSp equivalent to FSp. The 33-message history was organized as a combination of a \textit{fixed} and a \textit{sliding window}. We set the fixed window to ensure the model is always aware of the task we asked it to answer in the early prompts; instead, we set the sliding window to emulate the flow of the conversation as in ChatGPT Web. In particular, i) in ZSp, the fixed window covers our first prompt (i.e., task explanation) and the (Chat)GPT answer, while the sliding window covers the $i$-th prompts and the last 30 messages (i.e., 15 prompts and 15 (Chat)GPT answers); ii) in FSp, the fixed window covers the first 26 messages (i.e., task explanation and example instances), while the sliding window covers the i-th prompts and the last 6 messages. Figure~\ref{fig:prompt-api} summarizes the message history we set for testing GPT API.

\clearpage
\begin{minipage}{0.9\textwidth}
\centering
\includegraphics[width=0.95\textwidth]{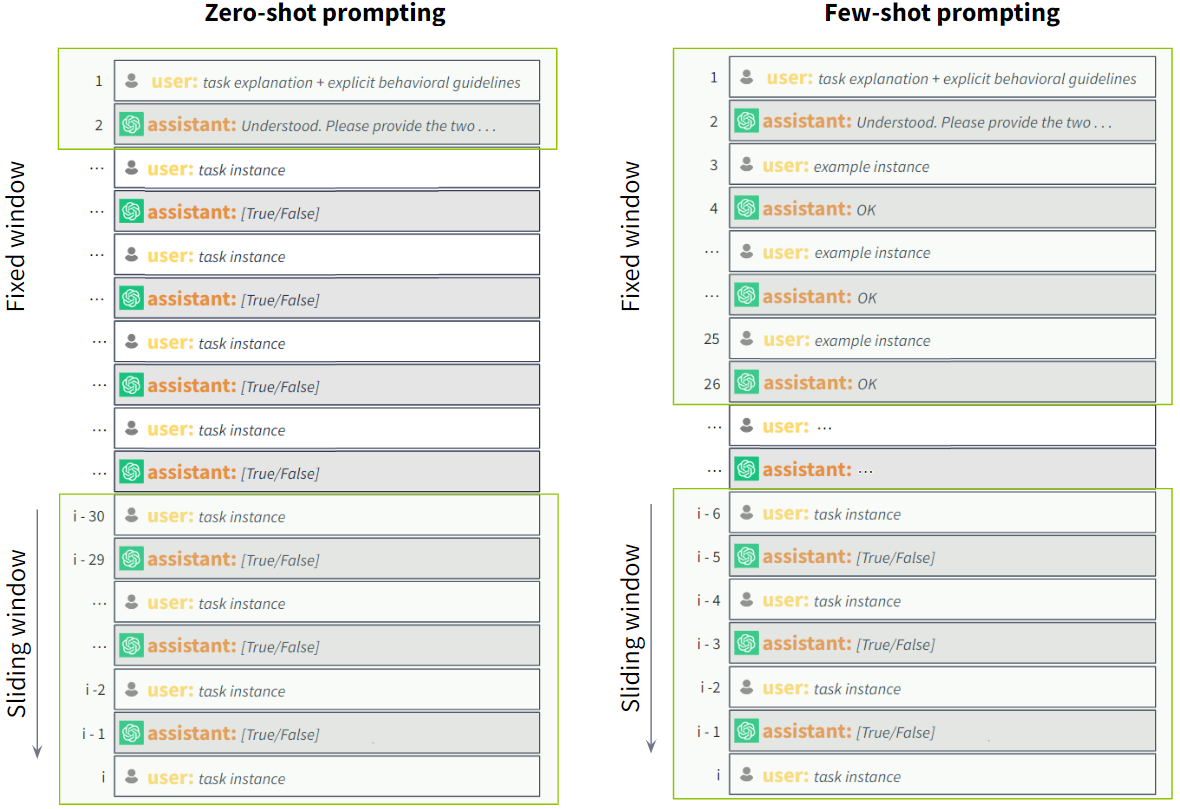}
\captionof{figure}{Message history used for GPT API in the zero-shot prompting (ZSp) and few-shot prompting (FSp) strategies. The message history is organized as a combination of a fixed and a sliding window, encompassing a total of 33 messages. The fixed window ensures that the model remains constantly aware of the task we have asked it to address in the initial prompts and the given examples (if any). Conversely, we establish the sliding window to emulate the conversational flow of ChatGPT Web.}
\label{fig:prompt-api}
\end{minipage}
\clearpage

\section{(Chat)GPT templates}\label{app:template}
\subsection{WiC template}\label{app:WiCtemplate}

\begin{minipage}{\textwidth}
\centering
\resizebox{\textwidth}{!}{%
\begin{tabular}{cl}
\textbf{Description} & \multicolumn{1}{c}{\textbf{Template}} \\ \hline
\multicolumn{1}{|c}{task explanation} & \multicolumn{1}{l|}{\begin{tabular}[c]{@{}l@{}}\textbf{Task}: Determine whether two given sentences use a target word with the same meaning or different meanings \\ in their respective contexts.\end{tabular}} \\ \hline
\multicolumn{1}{|c}{\begin{tabular}[c]{@{}c@{}}explicit behavioral \\ guidelines\end{tabular}} & \multicolumn{1}{l|}{\begin{tabular}[c]{@{}l@{}}I'll provide some negative and positive examples to teach you how to deal with the task before testing you.\\ Please respond with only "OK" during the examples; when it's your turn, answer only with "True" or "False"\\ without any additional text. When it's your turn, choose one: "True" if the target word has the same meaning in\\ both sentences; "False" if the target word has different meanings in the sentences. I'll notify you when it's your\\ turn.\end{tabular}} \\ \hline
\multicolumn{1}{|c}{example instance} & \multicolumn{1}{l|}{\begin{tabular}[c]{@{}l@{}}This is an example. You have to answer "OK":\\ \textbf{Sentence 1}: [First sentence containing the target word]\\ \textbf{Sentence 1}: [First sentence containing the target word]\\ \textbf{Target}: [Target word]\\ \textbf{Question}: Do the target word in both sentences have the same meaning in their respective contexts?\\ \textbf{Answer}: [True/False]\end{tabular}} \\ \hline
\multicolumn{1}{|c}{task instance} & \multicolumn{1}{l|}{\begin{tabular}[c]{@{}l@{}}Now it's your turn. You have to answer with "True" or "False":\\ \textbf{Sentence 1}: [First sentence containing the target word]\\ \textbf{Sentence 1}: [First sentence containing the target word]\\ \textbf{Target}: [Target word]\\ \textbf{Question}: Do the target word in both sentences have the same meaning in their respective contexts?\\ \textbf{Answer}: [The model is expected to respond with "True" or "False"]\end{tabular}} \\ \hline
\end{tabular}%
}
\captionof{table}{Sections of the prompt template used for testing (Chat)GPT.}
\label{tab:prompt-sections}
\end{minipage}

\vspace{0.1\textwidth}

\begin{minipage}{\textwidth}
\centering
\resizebox{0.55\columnwidth}{!}{%
\begin{tabular}{ccc}
\textbf{ID} & \textbf{Strategy} & \textbf{Prompt} \\ \hline
\multicolumn{1}{|c}{ZSp} & zero-shot prompting & \multicolumn{1}{c|}{\begin{tabular}[c]{@{}c@{}}task explanation\\ explicit behavioral guidelines\\ task instance\\ ...\\ task instance\end{tabular}} \\ \hline
\multicolumn{1}{|c}{FSp} & few-shot prompting & \multicolumn{1}{c|}{\begin{tabular}[c]{@{}c@{}}task explanation\\ explicit behavioral guidelines\\ example instance\\ ...\\ example instance\\ task instance\\ ...\\ task instance\end{tabular}} \\ \hline
\multicolumn{1}{|c}{MSp} & many-shot prompting & \multicolumn{1}{c|}{\textit{like FSp}} \\ \hline
\end{tabular}%
}
\captionof{table}{Prompt template for each employed prompting strategy.}
\label{tab:prompt-strategies}
\end{minipage}

\subsection{LSC template}\label{app:LSCtemplate}
\begin{minipage}{\textwidth}
\centering
\resizebox{\columnwidth}{!}{%
\begin{tabular}{cl}
\textbf{Strategy} & \multicolumn{1}{c}{\textbf{Template}} \\ \hline
\multicolumn{1}{|c}{ZSp} & \multicolumn{1}{l|}{\begin{tabular}[c]{@{}l@{}}Consider the following two time periods and target word. How much has the meaning of the target word \\ changed between the two periods? Rate the lexical semantic change on a scale from 0 to 1. Provide only a score.\\ \textbf{Target}: [Target word]\\ \textbf{Time period 1}: 1810--1860\\ \textbf{Time period 2}: 1960--2010\\ \textbf{Answer}: [The model is expected to respond with a continuous score $s$, with 0 $\leq s \leq$ 1 ]\end{tabular}} \\ \hline
\end{tabular}%
}
\captionof{table}{Prompt template for LSC.}
\label{tab:lsc}
\end{minipage}
\clearpage

\section{GPT API performance on TempoWiC and HistoWiC}\label{app:temperature}

\subsection{Experiment 1 - temperature}
\begin{minipage}{\textwidth}
\centering
\resizebox{0.9\textwidth}{!}{%
\begin{tabular}{cccccccccccccc}
 &  & \multicolumn{11}{c}{\textbf{GPT API - Temperature}} &  \\ \cline{3-13}
 & \multicolumn{1}{c|}{\textbf{prompt}} & \textit{0.0} & \textit{0.2} & \textit{0.4} & \textit{0.6} & \textit{0.8} & \textit{1.0} & \textit{1.2} & \textit{1.4} & \textit{1.6} & \textit{1.8} & \multicolumn{1}{c|}{\textit{2.0}} & \textbf{avg} \\ \cline{2-14} 
\multicolumn{1}{c|}{\multirow{2}{*}{\textbf{TempoWiC}}} & \multicolumn{1}{c|}{\textit{ZSp}} & .568 & .584 & .604 & .599 & .592 & .576 & .604 & .560 & .560 & .599 & \multicolumn{1}{c|}{.579} & \multicolumn{1}{c|}{.584} \\
\multicolumn{1}{c|}{} & \multicolumn{1}{c|}{\textit{FSp}} & .648 & .648 & .664 & .634 & .597 & .631 & .645 & .585 & .608 & .581 & \multicolumn{1}{c|}{.598} & \multicolumn{1}{c|}{.622} \\
\multicolumn{1}{c|}{\multirow{2}{*}{\textbf{HistoWiC}}} & \multicolumn{1}{c|}{\textit{ZSp}} & .728 & .683 & .689 & .676 & .666 & .694 & .715 & .609 & .704 & .671 & \multicolumn{1}{c|}{.594} & \multicolumn{1}{c|}{.675} \\
\multicolumn{1}{c|}{} & \multicolumn{1}{c|}{\textit{FSp}} & .684 & .698 & .721 & .698 & .671 & .700 & .686 & .599 & .552 & .607 & \multicolumn{1}{c|}{.601} & \multicolumn{1}{c|}{.656} \\ \cline{2-14} 
\end{tabular}%
}
\captionof{table}{Comparison of GPT performance (Macro-F1) for TempoWiC and HistoWiC at various temperature values using the official API and different prompts. }
\end{minipage}

\subsection{Experiment 2 - temperature}
\begin{minipage}{\textwidth}
\centering
\resizebox{0.9\textwidth}{!}{%
\begin{tabular}{cccccccccccccc}
 &  & \multicolumn{11}{c}{\textbf{GPT API - Temperature}} &  \\ \cline{3-13}
 & \multicolumn{1}{c|}{\textbf{prompt}} & \textit{0.0} & \textit{0.2} & \textit{0.4} & \textit{0.6} & \textit{0.8} & \textit{1.0} & \textit{1.2} & \textit{1.4} & \textit{1.6} & \textit{1.8} & \multicolumn{1}{c|}{\textit{2.0}} & \textbf{avg} \\ \cline{2-14} 
\multicolumn{1}{c|}{\multirow{2}{*}{\textbf{TempoWiC}}} & \multicolumn{1}{c|}{\textit{ZSp}} & .645 & .628 & .643 & .605 & .664 & .602 & .600 & .598 & .575 & .580 & \multicolumn{1}{c|}{.636} & \multicolumn{1}{c|}{.616} \\
\multicolumn{1}{c|}{} & \multicolumn{1}{c|}{\textit{FSp}} & .659 & .632 & .649 & .627 & .644 & .597 & .689 & .627 & .597 & .551 & \multicolumn{1}{c|}{.562} & \multicolumn{1}{c|}{.621} \\
\multicolumn{1}{c|}{\multirow{2}{*}{\textbf{HistoWiC}}} & \multicolumn{1}{c|}{\textit{ZSp}} & .751 & .758 & .711 & .765 & .729 & .712 & .678 & .652 & .679 & .664 & \multicolumn{1}{c|}{.604} & \multicolumn{1}{c|}{.700} \\
\multicolumn{1}{c|}{} & \multicolumn{1}{c|}{\textit{FSp}} & .684 & .678 & .707 & .700 & .706 & .665 & .607 & .662 & .615 & .592 & \multicolumn{1}{c|}{.623} & \multicolumn{1}{c|}{.658} \\ \cline{2-14} 
\end{tabular}%
}
\captionof{table}{Comparison of GPT performance (Macro-F1) for TempoWiC and HistoWiC at various temperature values using the official API and different prompts. }
\end{minipage}

\subsection{Average performance per temperature}
\begin{minipage}{\textwidth}
\centering
\resizebox{0.9\textwidth}{!}{%
\begin{tabular}{cccccccccccccc}
 &  & \multicolumn{11}{c}{\textbf{GPT API - Temperature}} &  \\ \cline{3-13}
 & \multicolumn{1}{c|}{\textbf{prompt}} & \textit{0.0} & \textit{0.2} & \textit{0.4} & \textit{0.6} & \textit{0.8} & \textit{1.0} & \textit{1.2} & \textit{1.4} & \textit{1.6} & \textit{1.8} & \multicolumn{1}{c|}{\textit{2.0}} & \textbf{avg} \\ \cline{2-14} 
\multicolumn{1}{c|}{\multirow{2}{*}{\textbf{TempoWiC}}} & \multicolumn{1}{c|}{\textit{ZSp}} & .606 & .606 & .624 & .602 & .628 & .589 & .602 & .579 & .568 & .589 & \multicolumn{1}{c|}{.607} & \multicolumn{1}{c|}{.600} \\
\multicolumn{1}{c|}{} & \multicolumn{1}{c|}{\textit{FSp}} & .654 & .640 & .657 & .631 & .620 & .614 & .667 & .606 & .602 & .566 & \multicolumn{1}{c|}{.580} & \multicolumn{1}{c|}{.622} \\
\multicolumn{1}{c|}{\multirow{2}{*}{\textbf{HistoWiC}}} & \multicolumn{1}{c|}{\textit{ZSp}} & .740 & .720 & .700 & .720 & .698 & .703 & .696 & .631 & .692 & .668 & \multicolumn{1}{c|}{.599} & \multicolumn{1}{c|}{.688} \\
\multicolumn{1}{c|}{} & \multicolumn{1}{c|}{\textit{FSp}} & .684 & .688 & .714 & .699 & .688 & .682 & .647 & .631 & .584 & .599 & \multicolumn{1}{c|}{.612} & \multicolumn{1}{c|}{.657} \\ \cline{2-14} 
\end{tabular}%
}
\captionof{table}{Comparison of GPT performance (Macro-F1) for TempoWiC and HistoWiC at various temperature values using the official API and different prompts. We report the average performance for each temperature.}
\label{tab:temperature}
\end{minipage}

\section{BERT performance on TempoWiC and HistoWiC}\label{app:bert}

\begin{minipage}{\textwidth}
\centering
\resizebox{0.8\textwidth}{!}{%
\begin{tabular}{cccccccccccccc}
                              & \multicolumn{13}{c}{\textbf{Layers}}                                                                                                     \\ \cline{2-14} 
\multicolumn{1}{c|}{\textit{}} &
  \textit{1} &
  \textit{2} &
  \textit{3} &
  \textit{4} &
  \textit{5} &
  \textit{6} &
  \textit{7} &
  \textit{8} &
  \textit{9} &
  \textit{10} &
  \textit{11} &
  \textit{12} &
  \multicolumn{1}{c|}{\textbf{avg}} \\ \cline{2-14} 
\multicolumn{1}{c|}{\textbf{TempoWiC}} & .669 & .631 & .635 & .627 & .604          & .627 & .704 & .749 & .744 & .730 & .737 & .751 & \multicolumn{1}{c|}{.684} \\ \cline{2-14} 
\multicolumn{1}{c|}{\textbf{HistoWiC}} & .650 & .678 & .739 & .782 & .828 & .801 & .806 & .771          & .771          & .749 & .722 & .744 & \multicolumn{1}{c|}{.753} \\ \cline{2-14} 
\end{tabular}}
\captionof{table}{Comparison of BERT Performance (Macro-F1) for TempoWiC and HistoWiC tasks at different embedding layers.}
\label{tab:bert}
\end{minipage}

\section{GPT API performance on LSC}\label{app:lsc}
\begin{minipage}{\textwidth}
\centering
\resizebox{0.8\textwidth}{!}{%
\begin{tabular}{ccllllllllll}
\textbf{} & \multicolumn{11}{c}{\textbf{Temperature}} \\ \cline{2-12} 
\multicolumn{1}{c|}{\textbf{}} & \textit{0.0} & \multicolumn{1}{c}{\textit{0.2}} & \multicolumn{1}{c}{\textit{0.4}} & \multicolumn{1}{c}{\textit{0.6}} & \multicolumn{1}{c}{\textit{0.8}} & \multicolumn{1}{c}{\textit{1.0}} & \multicolumn{1}{c}{\textit{1.2}} & \multicolumn{1}{c}{\textit{1.4}} & \multicolumn{1}{c}{\textit{1.6}} & \multicolumn{1}{c}{\textit{1.8}} & \multicolumn{1}{c|}{\textit{2.0}} \\ \cline{2-12} 
\multicolumn{1}{c|}{\textbf{SemEval-English}} & \multicolumn{1}{l}{.251} & .200 & .207 & .279 & .008 & .012 & .230 & .154 & .011 & .194 & \multicolumn{1}{l|}{.004} \\ \cline{2-12} 
\end{tabular}}
\captionof{table}{Comparison of (Chat)GPT performance (Spearman’s correlation) for LSC on SemEval-English at various temperature values using the official API.}
\label{tab:lsc-correlation}
\end{minipage}

\end{document}